\documentclass[11pt,letterpaper]{article}
\usepackage{emnlp2017}
\pdfoutput=1
\usepackage{times}
\usepackage{url}
\usepackage{latexsym}

\usepackage{amsmath}
\usepackage{graphicx}
\usepackage{color}
\usepackage{multirow}
\usepackage[utf8]{inputenc}
\usepackage{algorithm}
\usepackage{algorithmicx}
\usepackage{algpseudocode}
\usepackage{pifont}% http://ctan.org/pkg/pifont

\def\Tref#1{Table~\ref{#1}}
\def\Fref#1{Figure~\ref{#1}}

\def\footurl#1{\footnote{\url{#1}}}

\emnlpfinalcopy

\def\emnlppaperid{69}

 \author{Ale{\v{s}} Tamchyna$^{1,2}$ \and Marion Weller-Di Marco$^{1,3}$ \and Alexander Fraser$^1$  \\
         $^1$LMU Munich, $^2$Memsource, $^3$University of Stuttgart\\
         {\tt ales.tamchyna@memsource.com dimarco@ims.uni-stuttgart.de}\\ {\tt fraser@cis.lmu.de} } 

\begin{document}

\title{Modeling Target-Side Inflection in Neural Machine Translation}

\maketitle

\begin{abstract}

NMT systems have problems with large vocabulary sizes. Byte-pair
encoding (BPE) is a popular approach to solving this problem, but while BPE
allows the system to generate any target-side word, it does not enable
effective generalization
over the rich vocabulary in morphologically rich languages with strong
inflectional phenomena.
We introduce a simple approach to overcome this problem
by training a system to produce the lemma of a word and its morphologically 
rich POS tag, which is then followed by a deterministic generation step. 
We apply this strategy for English--Czech and English--German translation 
scenarios, obtaining improvements in both settings.
We furthermore show that the improvement is not due to only adding 
 explicit morphological information.
\end{abstract}

%
%%%%%%%%%%%%%%%%%%%%%%%%%%%%%%%%%%%%%%%%%%%%%%%%%%%%%%%%%%%%%%%%%%%%%%%%%%%%%%%%
%

\section{Introduction}

Neural machine translation (NMT) has recently become the new state of the art.
Despite a large body of recent research, NMT still remains a relatively
unexplored territory.

In this work, we focus on one of these less studied areas, namely target-side
morphology. NMT systems typically produce outputs word-by-word and at each step,
they evaluate the probability of all possible target words. When translating to
morphologically rich languages, due to the large size of target-side
vocabularies, NMT systems run into scalability issues and struggle with
vocabulary coverage.

Byte-pair encoding (BPE, \newcite{bpe}) is currently perhaps the most successful
approach to addressing these problems. However, while BPE allows the system to
generate any target-side word (possibly as a concatenation of smaller segments),
it does not enable effective generalization over the many different surface
forms possible for a single lemma, which had been shown to be useful in
phrase-based SMT \cite{2010-failures}. 

We see three main problems associated with rich target-side morphology in NMT:
(i) NMT systems have no explicit connection between different surface forms of a
single target-side lexeme (lemma), leading to data sparsity, (ii) there is no
explicit information about morphological features of target-side words, and
(iii) NMT systems cannot systematically generate unseen surface forms of known
lemmas: while the combination of subword segments obtained with BPE splitting
can technically generate new forms, this is not a linguistically informed way to
generate new words, and is furthermore restricted to ``simple'' concatenative
word formation processes. 

We propose a simple two-step approach to achieve morphological generalization in
NMT. In the first step, we use an encoder-decoder NMT system with attention and
BPE \cite{bahdanau, bpe} to generate a sequence of interleaving morphological
tags and lemmas. In the second step, we use a morphological generator to produce
the final inflected output. This decomposition addresses all three of the
problems outlined above:

\begin{itemize}
  \item the presence of lemmas allows the system to model different inflections
    jointly and better capture lexical correspondence with the source,
  \item morphological information is explicit and allows the system to easily
    learn target-side morpho-syntactic patterns including agreement,
  \item unseen surface forms can be generated simply by combining a known
    lemma and a known tag.  
\end{itemize}        

While simple, the approach is very effective and leads to significant
improvements in translation quality in a medium-resource setting for
English-Czech translation. Similarly, experiments in an English--German setting 
lead to improved translation results and also show that the proposed strategy can 
be applied to other language pairs.

%--------------------------------------------------------------------------------

\begin{figure*}[t!]
  \small
  \centering

  \renewcommand{\arraystretch}{1.07}
  \setlength{\tabcolsep}{0.4em}
  \begin{tabular}{rl}
    \bf input:     & there are a million different kinds of pizza . \\ 
    \hline
    \bf baseline:  & existují miliony druhů piz@@ zy . \\
    \bf morphgen: & VB-P---3P-AA--- existovat NNIP1-----A---- milión NNIP2-----A---- druh NNFS2-----A---- pizza Z:------------- . \\
  \end{tabular}
  \caption{Examples of input and output training sequences for the baseline and
  the proposed system. BPE splits are denoted by ``@@''.}
  \label{fig:input-output}
\end{figure*}

%--------------------------------------------------------------------------------

\begin{figure}[t!]
  \begin{center}
    \begin{tabular}{r|ll}
      \bf Category & \bf Value & \bf Description \\
      \hline
      POS        & A & adjective \\
      sub-POS    & A & adjective, general \\
      gender     & I & masculine inanimate \\
      number     & P & plural \\
      case       & 7 & instrumental \\
      possgender & -- & \it (possessor's gender) \\
      possnumber & -- & \it (possessor's number) \\
      person     & -- & \it (person, verbs) \\
      tense      & -- & \it (tense, verbs) \\
      grade      & 2  & comparative degree \\
      negation   & A  & affirmative (not negated) \\
      voice      & -- & \it (voice, verbs) \\
      reserve1   & -- & \it (unused) \\
      reserve2   & -- & \it (unused) \\
      var        & -- & \it (style, variant) \\
    \end{tabular}
    \caption{Czech positional tagset. Feature values for the word {\it kulatějšími}, tag {\tt AAIP7----2A----}.}
    \label{fig:czech-tagset}
  \end{center}
\end{figure}

%
%%%%%%%%%%%%%%%%%%%%%%%%%%%%%%%%%%%%%%%%%%%%%%%%%%%%%%%%%%%%%%%%%%%%%%%%%%%%%%%%
%

\section{Two-Step NMT}

\label{sec:two-step}

We work within the standard encoder-decoder framework with an attention
mechanism as proposed by \newcite{bahdanau}, using the Nematus implementation
\cite{nematus}.  
To model target-side morphology, the system is trained on an intermediate representation 
consisting of interleaved lemmas and morphological tags providing the full set of relevant 
inflection features.
Decoding is followed by a second step which is fully deterministic. We use the
predicted pairs of (tag+features, lemma) as input to a morphological generator which
outputs the final inflected surface forms. In the rare cases where the generator
fails to output
any surface
form, we simply output the lemma.

Our approach is inspired by the successful results of \newcite{syntax-nmt},
where the authors interleave target-side words
and CCG supertags and observe
improvements by learning to also predict the target-side syntax.
Our experiments in the English--Czech translation task will, however, show that the improvement
we obtain is not a similar effect, but instead
requires
the improved generalization obtained through mapping inflected forms to their 
lemmas and the ability to generate correct surface forms.

In this paper, we first apply our tag lemma strategy
to an English--Czech translation setting. We show that it is effective
and also investigate
potential effects of tag prediction interacting with morphological 
generalization.
A second set of experiments concerns English--German translation: here, the focus is rather
put on modeling linguistic phenomena, including German word formation. 
While Czech has a more complex morphology than German, German has the additional problem of
compounds that make translation challenging; one system variant thus includes simple compound handling.

%
%%%%%%%%%%%%%%%%%%%%%%%%%%%%%%%%%%%%%%%%%%%%%%%%%%%%%%%%%%%%%%%%%%%%%%%%%%%%%%%%
%

\section{Modeling Czech Morphology}

Czech is a Slavic language with a rich inflectional morphology. There are seven
cases for nouns and adjectives, four genders and two grammatical numbers.
Surface forms of verbs follow complex rules as well, as they encode number,
person, tense and several other phenomena. Due to its fusional nature, there is
a degree of syncretism in Czech -- words with different morphological features
may share the same surface form.

As such, Czech is a suitable example for evaluating our approach. We use the
Czech positional tagset in our work \cite{czech-tags}. 
\Fref{fig:input-output} illustrates the input and output to our network and the
baseline.
\Fref{fig:czech-tagset}
illustrates the tagset on an example. For Czech morphological analysis, tagging
and generation, we use the MorphoDiTa toolkit \cite{morphodita}, which achieves
state-of-the-art results in lemmatization and tagging and its coverage in
morphological generation is very high. Morphological generation is based on a
lexicon of lemmas and their paradigms and it is fully deterministic.

% 
%
%%%%%%%%%%%%%%%%%%%%%%%%%%%%%%%%%%%%%%%%%%%%%%%%%%%%%%%%%%%%%%%%%%%%%%%%%%%%%%%%
%

\section{Modeling German Morphology}

To obtain the representation of interleaved lemmas and tag+feature sequences for German,
we apply a slightly different pipeline than for the English--Czech setting.
Instead of representing a word by a simple lemma and a morphological tag, we use a morphological 
analyzer covering also productive formation processes -- the morphologically complex analyses  
of the lemma (``stem'') allow
us
to easily handle compounds, which pose a considerable challenge 
when translating into German.

\subsection{Linguistic Resources}
The key linguistic knowledge sources to model German morphology are the constituency parser
BitPar \cite{schmid04:coling} to obtain morphological analyses in the sentence context, and the morphological
tool SMOR \cite{schmid04:lrec} to analyze and generate inflected German surface forms.

SMOR is a a morphological analyzer for German inflection and word formation processes implemented in finite
state technology. In particular, it also covers productive word formation processes such as compounding or
derivation.
SMOR functions in two directions: {\it surface form $\rightarrow$ stem+features} and {\it stem+features 
$\rightarrow$ surface form}.
Thus, when preparing the target-side training data, each inflected surface form is analyzed, and then replaced 
by its stem and respective morphological features, as illustrated for the verb {\it trifft} below: \\[1ex]
\begin{tabular}{ll}
surface &{\it trifft} \\
stem & {\tt \small treffen<+V><3><Sg><Pres><Ind>}\\[1ex]
\end{tabular}

\noindent
For the inflection process after translation, SMOR is used in the reverse direction to output an inflected 
form when given a stem+feature sequence.

\subsection{German Inflectional Features}
German has a rich nominal and verbal morphology, and even though it exhibits a relatively high degree of 
syncretism, it has a high lemma--to--inflected forms ratio. For example, adjectives can have up to 6 different inflected forms, such as 
{\it blau, blaue, blaues, blauer, blauen, blauem} ('blue').

%---------------------------------------------------------------------------------
\begin{table*}[t]
\centering
\resizebox{\textwidth}{!}{
\begin{tabular}{|l|l|}
\hline
{\bf English}  & and what you 're seeing here is a cloud of densely packed , hydrogen-sulfide-rich water coming\\
{\bf sentence} & out of a volcanic axis on the sea floor  \\
\hline
\end{tabular}
}

\vspace*{4mm}
\resizebox{\textwidth}{!}{
\begin{tabular}{|l|l|l|c|l|}
\hline
{\bf EN gloss} & {\bf DE surface} & {\bf parse-tags} & {\bf infl.}&  {\bf fully specified stemmed representation}\\
\hline
{\it and} & und     & KON              & 0    & {\tt \small und[KON]} \\
\hline
{\it here}& hier    & ADV              & 1    & {\tt \small hier[ADV]} \\
\hline
{\it sees}& sieht   & VVFIN-Sg         & 1    & {\tt \small sehen$||$<+V><3><Sg><Pres><Ind>} \\
\hline
{\it one} & man     & PIS-Nom.Sg       & 0    & {\tt \small man[PIS]} \\
\hline
{\it a}   & eine    & ART-Acc.Sg.Fem   & 1    & {\tt \small eine<Indef>$||$<+ART><Fem><Acc><Sg><St>} \\
\hline
{\it cloud}& Wolke   & NN-Acc.Sg.Fem    & 1    & {\tt \small Wolke$||$<+NN><Fem><Acc><Sg><NA>} \\
\hline
{\it of}  & von     & APPR-Dat         & 0    & {\tt \small von[APPR-Dat]} \\
\hline
{\it dense}& dichtem & ADJA-Dat.Sg.Neut & 1    & {\tt \small dicht<Pos>$||$<+ADJ><Neut><Dat><Sg><St>} \\
\hline
{\it hydrogen-}& hydrogensulfid- &ADJA-Dat.Sg.Neut &1 & {\tt \small Hydrogen<NN>Sulfid<NN>} \\
{\it sulfide-rich}& reichem         &                 &  & {\tt \small reich<Pos>$||$<+ADJ><Neut><Dat><Sg><St>} \\
\hline
{\it water}& Wasser  & NN-Dat.Sg.Neut   & 1    & {\tt \small Wasser$||$<+NN><Neut><Dat><Sg><NA>} \\
\hline
{\it ,   }& ,       & \$,              & 0    & {\tt \small ,[\$]} \\
\hline
{\it that}& das     & PRELS-Nom.Sg.Neut& 0    & {\tt \small das[PRELS]} \\
\hline
{\it from}& aus     & APPR-Dat         & 0    & {\tt \small aus[APPR-Dat]} \\
\hline
{\it a}   & einer   & ART-Dat.Sg.Fem   & 1    & {\tt \small eine<Indef>$||$<+ART><Fem><Dat><Sg><St>} \\
\hline
{\it volcanic}& vulkanischen & ADJA-Dat.Sg.Fem & 1& {\tt \small vulkanisch$||$<+ADJ><Pos>}\\
               &              &                 &  &{\tt \small <NoGend><Dat><Sg><Wk>}\\ 
\hline
{\it longitudinal}& Längsachse   & NN-Dat.Sg.Fem   & 1& {\tt \small längs<ADJ>Achse$||$<+NN><Fem><Dat><Sg><NA>}\\
{\it axis}        &              &                 &  & \\
\hline
{\it on}  & an      & APPR-Dat         & 0    & {\tt \small an[APPR-Dat]} \\
\hline
{\it the} & dem     & ART-Dat.Sg.Masc  & 1    & {\tt \small die<Def>$||$<+ART><Masc><Dat><Sg><St>} \\
\hline
{\it sea floor}& Meeresboden & NN-Dat.Sg.Masc & 1  & {\tt \small Meer<NN>Boden$||$<+NN><Masc><Dat><Sg><NA>} \\
\hline
{\it oozes}& tritt   & VVFIN-Sg         & 1    & {\tt \small treten$||$<+V><3><Sg><Pres><Ind>} \\
\hline
{\tt .}& .       & \$.              & 0    & {\tt \small .[\$]} \\
\hline
\end{tabular}
}
\caption{Example for the fully specified representation used in the NMT system. The double-pipe symbol $||$
indicates the boundary between the word(stem) and the tag with the full set of inflectional features.}
\label{table:example-for-format}
\end{table*}
%---------------------------------------------------------------------------------

\paragraph{Nominal Inflection} ~
Unlike in English, where only the feature number is expressed for nouns, German nominal inflection is applied 
to determiners, adjectives and nouns. The following four features are relevant for nominal inflection:

\noindent
\resizebox{\columnwidth}{!}{
\begin{tabular}{ll}
case   & {\it nominative, accusative, dative, genitive} \\
gender & {\it feminine, masculine, neuter} \\
number & {\it singular, plural} \\
str/wk & {\it strong, weak}\\[1ex]
\end{tabular}
}

\noindent
To efficiently handle syncretism, SMOR has the artificial value {\it NoGend}, that is used when
a surface form is the same for all three values of gender; this is typical for plural forms.
Similarly, the feature strong/weak\footnote{Strong/weak inflection is determined by the setting of 
definite/indefinite articles in combination with the other feature: for example, the NP {\it das blau\textbf{e} 
Auto} ('the blue car') is inflected differently when occurring with an indefinite article ({\it ein blau\textbf{es} 
Auto}) in the function of subject or direct object.} 
does not need to be specified if the surface forms are the same;
we thus add the dummy-value {\tt \small <NA>} to always have a sequence of four values.
Words that are subject to nominal inflection are replaced by their SMOR analysis that is split 
into stem and the tag-feature sequence:

{\tt \small STEM <+Tag><Gend><Case><Num><St/Wk>}

\paragraph{Verbal Morphology} ~
German verbal morphology requires the modeling of these features:

\noindent
\begin{tabular}{ll}
person & {\it 1,2,3} \\
number & {\it singular, plural} \\
tense  & {\it present, past} \\
mood   & {\it indicative, subjunctive} \\[1ex] 
\end{tabular}

\noindent
These features refer to morphologically expressed properties in a single word; further instances of the
feature tense, in particular future tense, are realized as compound tenses.
Our modeling of verbal inflection, is restricted to the word-level, and the decision how to combine
auxiliaries and full verbs is left to the translation model. Verb forms are represented as follows in the
stemmed format:

\noindent
%\hspace*{-2mm}
\resizebox{1.03\columnwidth}{!}{
\begin{tabular}{ll}
finite & {\tt \small STEM <+V><Pers><Num><Tense><Mood>}\\
participle & {\tt \small STEM <+V><PPast>} \\
infinitive& {\tt \small STEM <+V><Inf>}\\
\end{tabular}
}

\subsection{Building the stemmed representation}

Table~\ref{table:example-for-format} illustrates the process of deriving the fully specified stemmed
representation by combining morphological analyses and rich parse tags; the column {\it infl} indicates
whether a word is inflected.
As a German surface form can have many possible analyses (cf. below), the parse tags are 
needed to disambiguate the morphological analyses.

\hspace*{-4mm}
\resizebox{\columnwidth}{!}{
\begin{tabular}{l}
{\it vulkanischen} \\
\hline
{\tt \small vulkanisch<+ADJ><Pos><Neut><Gen><Sg>}\\
{\tt \small vulkanisch<+ADJ><Pos><Masc><Acc><Sg>}\\
{\tt \small vulkanisch<+ADJ><Pos><Masc><Gen><Sg>}\\
{\tt \small vulkanisch<+ADJ><Pos><NoGend><Acc><Pl><Wk>}\\
{\tt \small vulkanisch<+ADJ><Pos><NoGend><Dat><Pl>}\\
\textbf{\texttt{\small vulkanisch<+ADJ><Pos><NoGend><Dat><Sg><Wk>}}\\
{\tt \small vulkanisch<+ADJ><Pos><NoGend><Gen><Pl><Wk>}\\
{\tt \small vulkanisch<+ADJ><Pos><NoGend><Nom><Pl><Wk>}\\
{\tt \small vulkanisch<+ADJ><Pos><Fem><Gen><Sg><Wk>}\\[1ex]
\end{tabular}
}

\noindent
The stem and the tag-feature sequence (or the bare tag for non-inflected words) are separated, 
allowing the model to learn lexical relations between source- and target-side separately from
target-side morpho-syntactic patterns. 
As the addition of tags effectively doubles the length of German sentences,  we also add tags (obtained with 
tree-tagger, \newcite{Schmid:94}) on the source-side to balance the source/target side sentence lengths.

\subsection{Reduction of Vocabulary Size}
One of the main objectives of the two-step approach is to reduce the target-side vocabulary size.
Table~\ref{table:most-frequent-surface-splits} 
shows the most frequent fragments on the end of words obtained through BPE splitting on the German 
surface data -- while it is difficult to generalize without the actual context, most tend to be inflectional 
suffixes. While this type of splitting does make sense, it also seems that there is some redundancy, and
a systematic generalization is impossible.
Furthermore, a mere segmentation of surface forms does not cover non-concatenative phenomena such as
``Umlautung'': for example, the concatenation of {\it Haus-} (lemma: 'house') and {\it -er} (typical plural 
suffix) does not result in the correct plural form ({\it Häuser}) -- thus, two ``lemmas'' are required to 
guarantee correct inflections of words that undergo Umlautung when working with surface forms. 
Table~\ref{table:comparison-vocabulary-size} shows the reduction of vocabulary in the stemmed representation: 
replacing inflected forms with their stems leads to a considerable reduction of the vocabulary size;
compound splitting leads to a further reduction.

%---------------------------------------------------------------------------------------
\begin{table}
%\small
\centering
\resizebox{\columnwidth}{!}{
\begin{tabular}{ll|ll|ll}
{\bf freq} & {\bf part} & {\bf freq} & {\bf part}& {\bf freq} & {\bf part} \\
\hline
 2469 & ten  & 1257 & sten   & 1077 & ern   \\
 2157 & te   & 1214 & es     & 1077 & -     \\
 1738 & en   & 1169 & ter    & 1058 & den   \\
 1607 & er   & 1148 & gen    & 1040 & s     \\
 1474 & ung  &1 078 & ischen & 1015 & ungen \\
\end{tabular}
}
\caption{The most frequent fragments on word ends after BPE from the German surface data.}
\label{table:most-frequent-surface-splits}
\end{table}
%all:
%less train.bpe.de | sed 's/\@\@ /@@/g' | tr ' ' '\n' | grep '@@' | sed 's/\@\@/ /g' | tr ' ' '\n' | sort | uniq -c | sort -nr | less 

% end of word:
%less train.bpe.de | sed 's/\@\@ /@@/g' | tr ' ' '\n' | grep '@@' | perl -e 'while(<>){chomp; s/\@\@/ @@/g; @line=split/ /; print "$line[-1]\n";}' | sort | uniq -c | sort -nr | less

%---------------------------------------------------------------------------------------
\begin{table}
\centering
\resizebox{\columnwidth}{!}{
\begin{tabular}{l|rr}
                      & {\bf vocabulary} & {\bf vocabulary} \\
                      & {\bf size}       & {\bf size w/ BPE} \\
\hline
{\bf DE surface data}       & 121.892     & 22.712   \\
{\bf DE morph}            & 97.587      & 21.663   \\
{\bf DE morph-split}      & 68.533      & 21.892   \\
\end{tabular}
}
\caption{Overview of vocabulary size in the German TED data (BPE: Byte Pair Encoding).}
\label{table:comparison-vocabulary-size}
\end{table}
%---------------------------------------------------------------------------------------

\subsection{Simple Compound Handling}
Another factor contributing to a high vocabulary size is the productivity of German compounds; in SMT,
compound handling has been found to improve translation quality, e.g. \newcite{stymnecancedda:2011} and
\newcite{cap:eacl2014}.
In addition to inflectional morphology, SMOR also provides a derivational analysis, including splitting into 
compound parts: for example, the compound {\it Häuser$|$markt} ('house market') is analyzed as 
{\tt \small Haus<NN>Markt<+NN><...>}.
In particular, the modifier is represented by its base form {\tt \small Haus}, covering the non-concatenative
process of ``Umlautung'' ({\it Haus $\leftrightarrow$ Häuser}).

In the stemmed representation, this may already present an indirect advantage, as compounds fragmented through BPE
splitting can match other stemmed occurrences of that word.
An obvious idea at this point is to go a step further and add compound splitting to the pre-processing
of the German data. Using the SMOR annotation, compounds are split at mid-word adjective and noun borders.
For example, the word {\it Meeres$|$boden} ('sea bottom') from table~\ref{table:example-for-format} is split into two sub-words
separated by the modifier's tag:

{\tt \small Meer §§<NN>§§ Boden <+NN><...>}

\noindent
This notation separates lexical parts  from SMOR markup, thus allowing the model to learn compound patterns.
After translation, the compound stems are concatenated and then inflected.

On the English side, it is assumed that the equivalents of compounds are already separate words.
For this system variant, however, the English side was slightly simplified by aggressive hyphen splitting,
and replacing nouns and verbs by their lemma form, accompanied by a tag indicating the type of inflection.
Our hope is that this representation will be more parallel to the compound-split representation in German.

\section{Experimental Evaluation}

In this section, we describe our experiments with English-Czech and
English-German translation.

\subsection{Czech}

We use the IWSLT training and test sets in English-Czech
experiments\footnote{\url{http://workshop2016.iwslt.org}}. The training set
consists of transcribed TED talks as collected in the WIT3 corpus \cite{wit3}.
We use IWSLT test set 2012 as the held-out set and the 2013 test set for
evaluation. \Tref{tab:data-sizes} summarizes the basic data statistics.

\begin{table}[t]
  \begin{center}
    \begin{tabular}{c|ccc}
      \bf corpus & \bf sents & \bf src tokens & \bf tgt tokens \\
      \hline
      train    & 114k  & 2309k & 1908k \\
      test2012 & 1385  & 25150 & 20682 \\
      test2013 & 1327  & 28454 & 24107 \\
    \end{tabular}
    \caption{Sizes of English-Czech corpora.}
    \label{tab:data-sizes}
  \end{center}
\end{table}

We use the Nematus toolkit for training the NMT systems \cite{nematus}. We run
BPE training on both sides of the training data with 49500 splits. We set the
vocabulary size to 50000 word types. The embedding size is set to 500, the
dimension of the hidden layer is 1024. We optimize the model using Adam
\cite{adam} and we use the default early stopping criterion in Nematus. We do
not apply drop-out anywhere in the model. Following \newcite{syntax-nmt}, we set
the maximum sequence length to 50 for the baseline and to 100 for systems which
produce interleaved outputs.

Our \emph{baseline} system is a standard Nematus setup with the parameters
described above. We refer to our two-step setup as \emph{morphgen} from now on.
For comparison, we also evaluate a third setting where we train the system to
output sequences of morphological tags interleaved with the surface forms. We
refer to this contrastive experiment as \emph{serialization} -- our aim is to tease
apart the possible benefit of explicitly predicting target-side morphological
tags from the improvements due to
morphological generalization.

Note that BPE is applied in all system variants. However, due to a reduced
vocabulary size in the \emph{morphgen} setting, the splits are uncommon and
morphological tags are never split (this is an effect of BPE, not a hard
constraint).

Because NMT system results can vary significantly due to randomness in
initialization and training, we run system training end-to-end for each variant
three times. We then select the best run based on BLEU as measured on the
development set (test2012) and then evaluate it on the final test set
(test2013).

\begin{table}[t]
  \begin{center}
    \begin{tabular}{c|cc}
      \bf system & \bf BLEU (dev) & \bf BLEU (test) \\
      \hline
      baseline   & 12.60 & 12.89 \\
      morphgen   & \bf 14.05 & \bf 14.57 \\
      serialization & 11.49 & 12.07 \\
    \end{tabular}
    \caption{English-Czech: BLEU scores of NMT system variants.}
    \label{tab:bleu}
  \end{center}
\end{table}

Importantly, the network was able to
learn the correct structure for both \emph{morphgen} and \emph{serialization}
systems. The outputs are well-formed sequences of interleaving tags and
lemmas/forms.

\Tref{tab:bleu} shows the obtained results. In our main experiment, our
two-step system achieves a substantial improvement of roughly 1.7 BLEU points,
showing that two-step in the neural context works for English to Czech
translation for this data size.

In the serialization experiment, we see that, surprisingly, the
\emph{serialization}
system does not outperform the baseline setup. This stands in contrast to the
use of CCG supertags by \newcite{syntax-nmt}, which was effective in this
framework. The result there showed that using CCG supertags which handle
syntactic generalization helps produce a better sequence of surface forms. We
attribute our result to the trade-off between providing the system with explicit
morpho-syntactic information (which is weaker information than CCG supertags)
and increasing the sequence length (which complicates training). It is possible
that with larger training data, \emph{serialization} might still outperform the
baseline, but our main result has shown that morphological generalization on
this data size is beneficial.

\paragraph{Scaling to Larger Data} ~ The observed improvements are certainly at
least partially due to reduced data sparsity: because Czech is a morphologically
rich language, there is a high number of distinct surface forms. We help the
system generalize by essentially dividing the information that surface forms
carry into two different ``streams'': one for morpho-syntax (tags) and the other
for semantics (lemmas).

One possible concern with the proposed approach is the ability to scale to
larger training data. Data sparsity could be such a major issue only when
training data are small and once we scale up, the observed benefits might
disappear as the system gets more robust statistical estimates for the
individual surface forms.

We run a targeted experiment with larger sizes of parallel training data to
determine whether the improvements hold. We always use the main training set
described above but additionally, we add a random sample from the CzEng 1.0
parallel corpus \cite{czeng10:lrec2012} to achieve training data sizes of 250
thousand up to 2 million parallel sentences (total).

\begin{table}[t]
  \begin{center}
    \begin{tabular}{r|ccc}
      & baseline & morphgen & $\Delta$ \\
      \hline
      IWSLT & 12.89 & 14.57 & 1.68 \\
      250k  & 14.87 & 17.51 & 2.64 \\
      500k  & 16.96 & 20.05 & {\bf 3.09} \\
      1M    & 18.07 & 20.95 & 2.88 \\
      2M    & 20.04 & {\bf 22.31} & 2.27 \\
    \end{tabular}
    \caption{English-Czech: BLEU scores of systems with larger parallel training data.}
    \label{tab:scaling-bleu}
  \end{center}
\end{table}

\Tref{tab:scaling-bleu} shows the results. We observe the highest difference in
the 500k setting (over 3 BLEU points absolute) and while the improvement
decreases slightly as we add more data, the difference is still around 2.3 BLEU
points even in the largest evaluated setting, which is an encouraging result.

Note that due to the increased computational cost, scores for larger system
variants are only based on a single training run.

\paragraph{Analysis and Discussion} ~ We now further analyze our two-step
system, \emph{morphgen}, in the IWSLT data setting. We first look at cases where
the generator failed to produce the surface form. We found only a handful of
cases; these mostly involved unknown proper names (Braper, Hvanda). In just four
cases, the tag proposed by the network was not compatible with the lemma (i.e.,
the network made an error).

\label{sec:analysis}

In order to determine where the improvement comes from, we analyze the number of
novel surface forms produced by the system. We find that indeed, unseen word
forms \emph{are} generated by the system but not nearly as many as we expected:
only 125 novel tokens were found in the test set (114 word types). Out of these,
14 forms are confirmed by the reference sentences (note that the unconfirmed
words may still be correct within the system output).

It seems that the system mostly benefits from the decomposition that we proposed
-- Czech lemmas are more easily mapped to source-side English words than the
many inflected forms associated with each lemma. The interleaving tags then help
explicitly train the morpho-syntactic structure of the sentences and allow the
second step to deterministically generate the final translations. While
morphological generalization does indeed occur, it is not the source of most of
the observed improvement.
When we use surface forms together with the annotations (in our serialization experiment), 
we see no improvement.

Finally, we report the results of a blind manual annotation contrasting outputs
of \emph{baseline} and \emph{morphgen}. For each instance, the annotator had
access to the reference translation and both outputs. The task was to rank which
translation is better or to mark both as equal quality. The annotator analyzed
200 sentences. In 130 cases, the translations were judged as equal. Out of the
remaining 70 sentences, the \emph{morphgen} system was marked as better in 48
cases and the baseline won in 22 cases.

\subsection{German}

The
initial
English--German experiments are evaluated on IWSLT training and test data, which consists of transcribed 
TED talks. The system is optimized on the 2012 dev-set (1165 sentences), and tested on the 2013 test-set (1363
sentences) and the 2014 test-set (1305 sentences). The training data consists of 184.879 parallel sentences,
after filtering out sentences shorter than 5 or longer than 50 words, as well as sentences that could not be 
parsed.
Prior to training the NMT system, the (stemmed) source- and target-data undergo BPE splitting (29500 splits), 
in order to keep the vocabulary within the predefined limit.

The translation experiments are carried out with the Nematus toolkit \cite{nematus}, using the training
parameters as displayed below, in combination with the default early stopping criterion in Nematus:

{\small
\begin{center}
\begin{tabular}{|ll|ll|}
\hline
vocab     & 30k     & dropout      & yes \\
dim\_word & 500     & dropout\_emb & 0.2 \\
dim       & 1024    & dropout\_hid & 0.2 \\
lrate     & 0.0001  & dropout\_src & 0.1 \\
opt       & adam    & dropout\_trg & 0.1 \\
maxlen    & 50(100) & & \\
\hline
\end{tabular}
\end{center}
}

\noindent
The sentence length is set to 50 for the baseline system, and extended to 100 for the morph-gen systems,
because the addition of the morphological tags doubles the sentence length.

\begin{table}
\centering
\begin{tabular}{l|rr|r}
  {\bf  TED'13 }  & {\bf run-1} &  {\bf run-2} & {\bf avg.} \\
\hline
{ baseline}  & 19.87   & 20.15 & 20.01\\
% \hline
{ morph-gen} & 20.73   & 20.98 & 20.86\\
% \hline
{ morph-gen-split} & 20.88 & 21.18 & 21.03\\
\end{tabular}

\vspace*{3mm}
\begin{tabular}{l|rr|r}
  {\bf  TED'14 }  & {\bf run-1} &  {\bf run-2} & {\bf avg.} \\
\hline
{ baseline}  & 19.02   & 18.68 & 18.85\\
% \hline
{ morph-gen} & 20.01   & 19.93 & 19.97\\
% \hline
{ morph-gen-split} & 20.07 & 20.76  & 20.42\\
\end{tabular}
\caption{English--German: lowercased BLEU for two test sests (1363 and 1305 sentences).}
\label{table:en-de-results}
\end{table}
%IWSLT16.TEDX.tst2013 -- ref is tokenized
%IWSLT16.TED.tst2014  -- ref is not tokenized
%slight improvements after fixing the selection of SMOR's generated compounds: Solarstrom vs. Solaresstrom

Table~\ref{table:en-de-results} shows the results for the English--German translation experiments, averaged over
two training runs:
on both test sets, the system generating inflected forms based on stems and features is better than the baseline.

Despite SMOR's complicated  structure, the resulting stems are generally well-formed; for un-inflectable stems
(mostly made-up words such as {\tt \small Parunelogramm<+NN><Neut><Gen><Sg>}), the markup is simply removed.

The addition of compound splitting leads to a minor further improvement.
We consider this a promising result, indicating that segmentation using the rich information provided by SMOR
can be helpful; we plan to explore this further in future work.

\begin{table}
\center
\resizebox{\columnwidth}{!}{
\begin{tabular}{l|ccc}
             & {\bf baseline} & {\bf morph-gen} & {\bf morph-gen-split}\\
\hline
 {\bf 250k } & 18.75 & 20.55 & 20.51\\
 {\bf 500k } & 21.39 & 22.79 & 23.00\\
\end{tabular}
}
\caption{English--German: lowercased BLEU for newstest'16 (2169 sentences) trained on 250k and 500k sentences news-mix data.}
\label{table:news-en-de}
\end{table}

\paragraph{Generation of novel words}
A closer look at the translation output reveals that there are indeed new word forms generated by the
{\it morph-gen} system. For the TED'13 set, for example, the {\it morph-gen} system output a total of 261 
words that are not in the training data or the English input sentence. Of these, 112 are names or nonsense words 
produced by concatenating BPE segments\footnote{Into this category, we also count non-wellformed generations 
by SMOR caused by incorrect transitional elements in compounds, e.g. {\it Oszillation\textbf{en}generator} vs. 
{\it Oszillation\textbf{s}generator}.}.
The other 149 words are morphologically well-formed, though not necessarily semantically sound (e.g.
{\it Schokoladenredakteur}: 'chocolate editor' as proposed translation for 'smart-ass editor') or appropriate
in the translation context. Thus, we compared the novel words with the reference translations: 23 words (21 nouns, 
2 adjectives) were found in the reference of the respective sentence. Of course, this under-estimates the number
of useful new creations, as a valid translation does not necessarily need to match exactly with the reference.
For the {\it morph-gen-split} system, only 27 matches with the reference were found in a set of
328 unseen forms.

\paragraph{Different Domain and Larger Corpus} ~ 
To assess the influence of domain and corpus size, we also evaluate the approach to model German
morphology in a larger news corpus setting.
To obtain a training corpus that is diverse, but still restricted in size, we combined randomly selected 
sentences (between 5-50 words) from the 4 parallel corpora provided for EN--DE translation at the WMT'17 
shared task\footnote{http://www.statmt.org/wmt17/translation-task.html} (selected in equal parts from Europarl, 
CommonCrawl, News-Commentary and RapidCorpus), resulting in a set of 250k and 500k sentences  
The model is optimized on newstest'15 and evaluated on newstest'16; table~\ref{table:news-en-de} shows
the results for the surface form baseline and the morphological generation systems with and without compound 
handling.
As for the TED data set, the morphological generation systems outperforms the systems trained on surface data,
but the improvement for the system trained on 500k sentences is slightly lower than for the system trained on 
250k sentences.
The systems with additional compound splitting obtained the same result as the basis morphological generation 
system (250k), or were slightly better (500k). With regard to the effectiveness of compound handling, it is 
difficult to draw a clear conclusion, but, looking also at the results obtained in the TED setting, it seems 
that there is a tendency that compound handling leads to a slight improvement.
As compounding is a productive word formation process that is challenging to cover even in large corpora,
compound handling might be useful also when using larger data training corpora.

%
%%%%%%%%%%%%%%%%%%%%%%%%%%%%%%%%%%%%%%%%%%%%%%%%%%%%%%%%%%%%%%%%%%%%%%%%%%%%%%%%
%

\section{Related Work}

\label{sec:related-work}

Generation
of unseen morphological variants has been tackled in various ways in
the context of phrase-based models and other SMT approaches. Notably, two-step SMT
was proposed to address this problem \cite{toutanova2008,2010-failures,twostep}.
In two-step SMT, a separate prediction model
(such as a linear-chain CRF)
is used to either directly predict
the surface form (as in \newcite{toutanova2008}) or used to predict the grammatical features,
following which morphological generation is performed (as in \newcite{2010-failures,twostep}).
Our work differs from
their work
in that we do not use a separate prediction model, but
instead rely on predicting the lemmas and surface-forms as a single sequence
in a neural machine translation model.

\newcite{huck17:eacl} recently proposed an approach related to two-step MT where
the unseen surface forms are added as synthetic phrases directly in the system
phrase table and a context-aware discriminative model is applied to score the
unseen variants. Unlike our work, the authors report diminishing improvements as
training data grows larger. Our approach learns a more robust underlying model
thanks to the reduced data sparsity. Unlike \newcite{huck17:eacl}, our
improvements are therefore not only due to the ability to generate words which
were not seen in the training data.

Factored translation models \cite{factoredmt} can deal with unseen word forms
thanks to generation steps. One of the original goals of factored MT was in fact
the scenario where the system produces lemmas and tags and then a generation
step could be used to produce the inflected forms. Factored models failed to
achieve this goal due to lemmas and tags being predicted independently, leading
to many invalid combinations, and due to the involved combinatorial explosion.

\newcite{martinez-factored-nmt} attempt to include target-side factors in neural
MT. Unlike our simple technique, their approach requires modifications to the
network architecture. The authors work with English-French translation and they
report mixed results.

Another successful attempt to learn novel inflections in SMT is back-translation
\cite{bojar-tamchyna:2011:WMT}. By using an MT system trained to translate
\emph{lemmas} in the opposite direction, it is possible to create synthetic
parallel data which contain unseen word forms of known lemmas on the target
side.  There are two main downsides to this approach. The first is that the
source language contains translation errors, which may affect translation
quality. The second is that the substitution of different surface forms for the
same target language lemma may result in incoherent translations, where the
context no longer agrees with the chosen surface form.
\newcite{nmt-backtranslation} propose to use back-translation in NMT to include
language modeling data, but the ``inverse'' NMT system is not able to translate
unseen target word forms (no lemmatization is done) and therefore this method
does not learn novel inflections. Applying BPE splitting can technically lead to
new inflected word forms, but this requires an appropriate segmentation into
base form and inflectional suffixes which might not always be the case, in
particular for infrequent words.

A very similar method to our two-step setting was independently proposed for use
in a natural language generation (NLG) pipeline for morphologically rich
languages \cite{odusek-phd}. However, in this scenario, the approach was
not better than a baseline which operated on surface forms.

Finally, there has been further more recent work on alternatives to
using BPE segmentation for NMT. \newcite{ataman17:eamt} looked at
segmentation for Turkish, which is an agglutinative
language. \newcite{huck17:wmt} presents an approach for segmenting
German with a focus on compound splitting and splitting suffixes off
of stems using a stemmer, which may allow generalization in a similar
way to our work. It would be interesting to compare with these approaches
in future work.

%
%%%%%%%%%%%%%%%%%%%%%%%%%%%%%%%%%%%%%%%%%%%%%%%%%%%%%%%%%%%%%%%%%%%%%%%%%%%%%%%%
%

\section{Conclusion}

\label{sec:conclusion}

In this work we showed that a simple setup, interspersing lemmas and rich
morphological tags, followed by deterministic generation of the resulting
surface form, results in impressive gains in NMT of English to Czech. Applying
the technique to an English to German system also resulted in considerable
improvements.
For English--German,
the addition of compound handling yielded promising results. Furthermore, among the novel
word forms for German, most were compounds -- as compounding is a very productive process, 
this is also a challenging problem when using larger corpora.
Exploring strategies for better segmentation and compound handling is an interesting task
that we plan to investigate further.

We believe that while simple, this technique effectively addresses the
fundamental problems of rich target-side morphology: (i) sparse data and lack of
connection between different forms of a single target lexeme, (ii) lack of
explicit morphological information, and (iii) inability to generate unseen forms
of known lexemes. Our results indicate that most of the improvement comes from the
first two properties.

Perhaps a modified training criterion could be used to encourage the system to
generalize more; in the standard setting, the system probably learns to strongly
condition the lemma on the tag and avoids the risk of generating new pairs. In
the situations where a novel form is required, the system may either bypass this
by producing a synonymous word or paraphrase, or it might simply produce an
ungrammatical form of the correct lemma. This phenomenon deserves more
examination which we leave to future work.

We further analyzed the serialization scenario, showing that the effect here is
not due to training the system to also predict morphological tags, which is in
contrast with the result of \newcite{syntax-nmt}. It is likely that the two
approaches are complementary, the rich information in CCG supertags could bring
additional benefit to the morphological generalization that we perform. We plan
to investigate this in future work.

%
%%%%%%%%%%%%%%%%%%%%%%%%%%%%%%%%%%%%%%%%%%%%%%%%%%%%%%%%%%%%%%%%%%%%%%%%%%%%%%%%
%

\section*{Acknowledgments}

This project has received funding from the European Union’s Horizon
2020 research and innovation programme under grant agreement No 644402
(HimL). This project has received funding from the European Research
Council (ERC) under the European Union's Horizon 2020 research and
innovation programme (grant agreement No 640550).

\bibliographystyle{emnlp_natbib} \bibliography{biblio}

\end{document}